\documentclass{llncs}
\usepackage{graphicx}
\usepackage{gensymb}
\usepackage[table]{xcolor}
\usepackage{makecell}
\usepackage{diagbox}
\usepackage{llncsdoc}
\begin{document}

%
\title{Endoscopic Depth Measurement and Super-Spectral-Resolution Imaging}

\author{Jianyu Lin\inst{1,2} \and Neil Clancy\inst{1,3} \and Yang Hu\inst{1,2} \and Ji Qi\inst{1,3} \and Taran Tatla\inst{4} \and Danail Stoyanov\inst{5,6} \and Lena Maier-Hein\inst{7} \and Daniel Elson\inst{1,3}}

\institute{
 \textsuperscript{1} Hamlyn Centre for Robotic Surgery, \inst{2} Dept. of Computing, \inst{3} Dept. of Surgery and Cancer, Imperial College London, London, SW7 2AZ, UK\\
\inst{4} Dept. of Otolaryngology - Head and Neck Surgery, Northwick Park Hospital, London, HA1 3UJ, UK\\
\inst{5} Centre for Medical Image Computing, \inst{6} Dept. of Computer Science, University College London, London, WC1E 7JE, UK\\
\inst{7} Div. of Computer Assisted Medical Interventions, German Cancer Research Center, Heidelberg, 69120, Germany}

\maketitle
\begin{abstract}
Intra-operative measurements of tissue shape and multi/ hyperspectral information have the potential to provide surgical guidance and decision making support. We report an optical probe based system to combine sparse hyperspectral measurements and spectrally-encoded structured lighting (SL) for surface measurements. The system provides informative signals for navigation with a surgical interface. By rapidly switching between SL and white light (WL) modes, SL information is combined with structure-from-motion (SfM) from white light images, based on SURF feature detection and Lucas-Kanade (LK) optical flow to provide quasi-dense surface shape reconstruction with known scale in real-time. Furthermore, “super-spectral-resolution” was realized, whereby the RGB images and sparse hyperspectral data were integrated to recover dense pixel-level hyperspectral stacks, by using convolutional neural networks to upscale the wavelength dimension. Validation and demonstration of this system is reported on \textit{ex vivo}/\textit{in vivo} animal/ human experiments.\\

\end{abstract}

\section{Introduction}
Using intra-operative information to aid surgical planning, navigation, and decision making is important for minimally invasive and robotic surgery. The data is mainly collected via endoscopes and other integrated hardware to provide real-time texture and color information. By data analysis the tissue surface shape can be extracted to register the intra- and pre-operative information from imaging modalities like CT and MRI \cite{Maier-Hein2013}.Intra-operative optical modalities such as multi/hyperspectral imaging (MSI/HSI) also have significant clinical impact, e.g. 1) narrow band imaging for vascular visualization; 2) oxygen saturation for intra-operative perfusion monitoring and clinical decision making; 3) tissue classification and pathology identification \cite{Lu2014,Clancy2015}.

Previously, a SL-enabled 3D tissue surface shape and hyperspectral imaging system was presented  \cite{Lin2015,Lin2016}. This used an optical fiber bundle with the fibers arranged in linear and circular arrays, respectively, at either end (Fig. \ref{fig:Fig1} (a)). In SL mode dispersed supercontinuum laser light was directed onto the linear array to emerge from the circular array as a spectrally encoded spot pattern. The tissue shape could be reconstructed sparsely if this light pattern was projected onto the surface and analyzed. Unlike passive stereo techniques, this SL reconstruction is not restricted by the texture information on the object surface. In HSI mode, the endoscopic white light illuminated the target surface. The reflected light was captured by the circular fiber array, emerged from the linear array, and imaged onto a slit HSI camera. The positions of detected spots in SL mode therefore indicated the locations of HSI signal on the white light images. However, the system did not provide surgeons with a WL view containing texture information, since no RGB images were captured. Further-more, it also suffered from sparse reconstructed surface and HSI signal, due to the finite number of fibers in the bundle. In this work, these two problems were addressed.

In this work a chopper wheel was used to stroboscopically switch between the SL and WL modes, to provide the surgeons with WL view. Both views were used to provide a quasi-dense reconstruction with known scale, using both SL and SfM \cite{Hartley2003}. This procedure was applied on a GPU to guarantee fast processing. Secondly, a deep learning-based method was studied to generate pixel-level dense multispectral image (MSI) stacks from RGB images and the sparse HSI signals.

HSI systems can be divided into two main types: spatial (e.g. with slit HSI camera) and spectral scanning systems (e.g. liquid crystal tunable filter (LCTF), or filter wheel multispectral cameras). However, there is always a trade-off between spatial resolution, spectral resolution and acquisition time, which affects surgical applications where the tissue is deformable and moving. Du \textit{et al.} proposed to use non-rigid registration to align mismatched HSI stacks \cite{Du2015}, but such methods are limited to relatively long, off-line processing. Recently, convolutional neural networks have been used to solve the image super-resolution problem, mainly to increase the spatial resolution of input images. Shi \textit{et al.} proposed fully convolutional networks to improve the image quality \cite{Shi2016}; Oktay \textit{et al.} applied residual networks to upscale low-resolution 3D MRI data \cite{Oktay2016}.In this paper, we developed a model to merge information from dense RGB images and sparse HSI signals to predict the corresponding dense 3D MSI stack with 24 channels. This was realized by upscaling the RGB images on its channel dimension and integrating sparse HSI signals to fine-tune the spectral shape on different locations. We refer to this method as “super-spectral-resolution”, i.e., achieving spatial super-resolution of sparse multispectral measurements by upscaling dense WL imag-es in the spectral domain. The proposed model was trained on \textit{ex vivo} and \textit{in vivo} in human and animal experiments.

In a nutshell, there are three key contributions in this work: 1) hardware improvement to provide WL views; 2) combination of SL and SfM for quasi-dense reconstruction; 3) real-time dense MSI using RGB images and sparse HSI signals.

\section{Materials and Methods}
\subsection{Interleaved SL and WL views}
\begin{figure}
\includegraphics[width=1\textwidth]{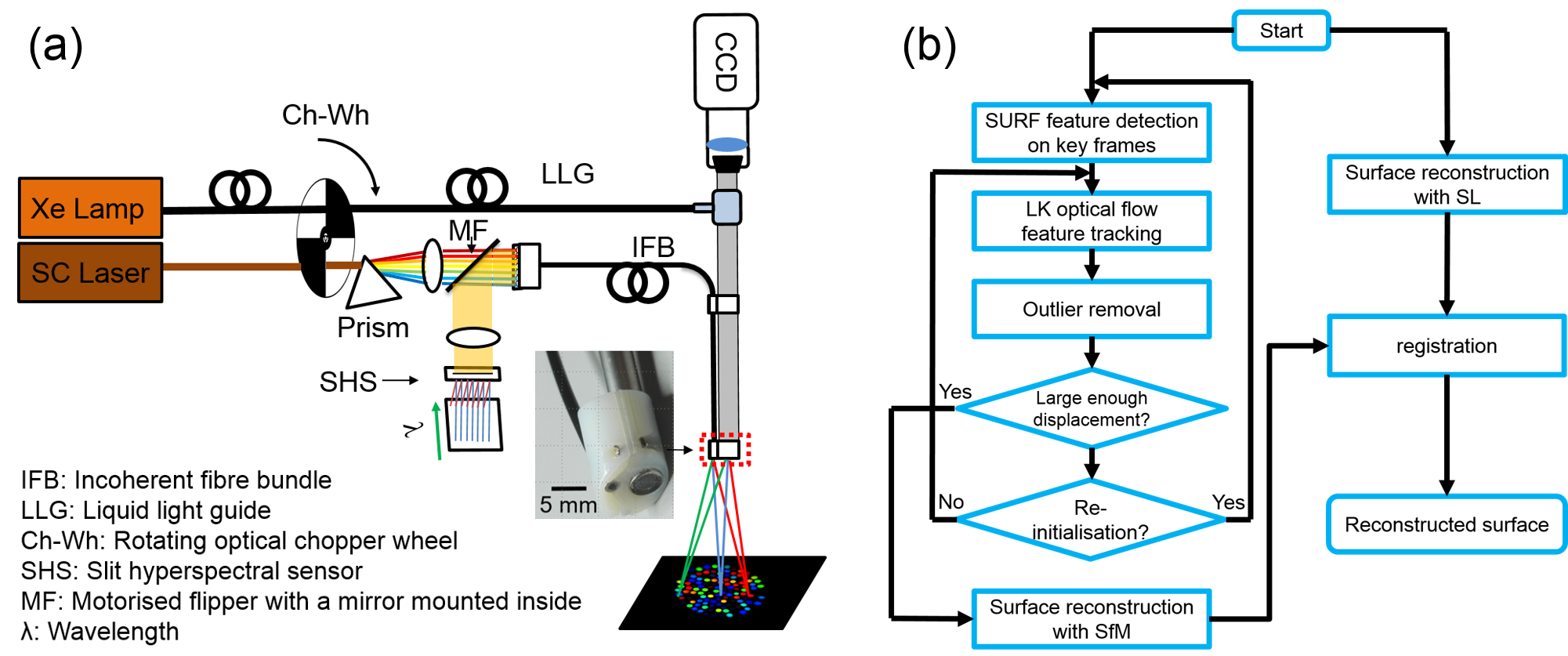}
\caption{(a) The schematic of the setup in SL mode, with a chopper wheel providing interleaved SL/WL views. (b) The tissue surface reconstruction workflow.}
\label{fig:Fig1}
\end{figure}

Rapid stroboscopic switching between WL and SL was achieved using an optical chopper wheel (3501 Optical Chopper; New Focus, Inc., USA) \cite{clancy2012} as shown in Fig.\ref{fig:Fig1} (a). Two fiber optic light cables were used for WL: one was connected to a xenon lamp and the other to the laparoscope. Their free ends were then positioned against each other, separated by a $2 mm$ air gap through which the chopper wheel could pass. The chopper was mounted so that the emitted supercontinuum laser also passed through the wheel, and the SL and WL beam paths were alternately blocked or transmitted as it turned. The result was that the light emerging from the tip of the instrument switched between SL and xenon at the chopping frequency. Separately, a computer-controlled signal generation device (NI USB-6211; National Instruments Corporation, USA) was used to produce two synchronized square waveforms of variable frequency and phase. One was used to trigger image acquisition by the CCD camera, while the other controlled the rotation frequency and the phase of the chop-per wheel. The trigger frequency was set to twice that of the chopper and the phase adjusted so that the acquired frames comprised of alternating SL and WL-illuminated images.

A tip adapter was 3D printed to mount the SL probe on a rigid endoscope ($5 mm$ diameter Hopkins II Optik 30\degree, Karl Storz GmbH, Germany). This adapter was cylindrical ($12 mm$ diameter), with two channels to house the endoscope and SL probe. The angle and baseline of these two channels were set to 10\degree and $5 mm$, to maximize triangulation accuracy for surface reconstruction within $~1.5$ -- $4 cm$ working distances.

\subsection{Tissue surface feature tracking and shape measurement}
In this work, we propose to combine information from both the SL and WL images. Previously, an SL reconstruction technique with fully convolutional networks (FCN) has been proposed, and worked robustly at a frame rate of ~12 FPS \cite{Lin2016}. However, due to \textit{in vivo} factors like strong light tissue interaction and CCD over-exposure, using SL alone does not always return a dense reconstruction. Therefore, we combined SL and SfM on a GPU to increase the reconstruction density and robustness (Fig. \ref{fig:Fig1} (b)).

\subsubsection{Surface reconstruction using monocular SfM.}
In this work, a method combining SURF-based feature detection and LK optical flow-based tracking has been proposed to perform a correspondence search. Several criteria were applied to exclude the tracking outliers, including the feature descriptor difference, flow vector length, temporal smoothness, symmetric optical flow, and RANSAC in essential matrix estimation. For the surface reconstruction we assumed the surface was rigid in a small time window. Then the relative position between the cameras in the two frames, as well as the up-to-scale 3D positions of the feature points, can be estimated using singular value decomposition (SVD) and examining all four possible solutions. Given enough correspondences, the surface can be measured using two adjacent frames.
\subsubsection{Combination of reconstruction results from SL and SfM.}
Since each WL image had two temporally adjacent SL images, the average shape reconstructed from two SL frames was used to register the SfM reconstruction results with scale information.

\subsection{Super-spectral-resolution imaging}
Super-resolution, recovering high-resolution (HR) images from their low resolution (LR) counterparts, is an ill-posed method, where one LR input could be mapped to multiple HR outputs. To solve this problem, two assumptions were made in this work: 1) The HR information is redundant in the HR images and could be partially extract-ed from the LR ones. 2) The mapping from LR to HR can be learnt from training sets containing data similar to the unseen data. The proposed approach upscaled the spectral dimension rather than the spatial dimension. Two models were developed: one recovers MSI stacks from RGB images only, while the other combines RGB images with the sparsely collected hyperspectral signal to further refine the MSI prediction.

\subsubsection{Model 1 - Recovering MSI stacks from RGB images.}
An RGB image was considered as an MSI stack with 3 spectral channels. The pro-posed model (Fig.\ref{fig:Fig2} (a)) looked for a mapping from an $M\times N\times 3$ MSI stack to $M\times N\times 24$, where $M\times N$ stands for the image spatial resolution. This model consists of two main stages:
\begin{itemize}
\item Upscaling the input in the spectral dimension. Four 3D transposed convolutional layers were piled together to transform the input from $M\times N\times 3$ to $M\times N\times 24$.
\item High-frequency-signal-extraction (HFE). This extracts and combines the high frequency signal with LR stacks. This stage was implemented using a residual block which introduces a “shortcut” to reduce the degradation of the training accuracy problem when deep networks are used. In our model the convolutional mapping $F(x)$ was used to extract the high frequency from input $x$, and then added to the "shortcut" $x$ input itself which represents a stack without high frequency content.
\end{itemize}
The structure of model 1 can be found in Fig.\ref{fig:Fig2} (a).This mapping achieved generally good spectral prediction but still with noticeable errors. To refine the predication, we extended model 1 to incorporate spatially sparse HSI signals captured using the system’s HSI mode.

\subsubsection{Model 2 - Recovering HSI stacks from RGB images and sparse spectral signals.}
RGB images provided high spatial but low spectral resolution; while HSI mode had low spatial but high spectral resolution. Due to the sparsity of the hyperspectral signal, RGB was used as the main contributor for MSI stack estimation, then the HSI signal was applied to refine the estimation.

\begin{figure}
\includegraphics[width=1\textwidth]{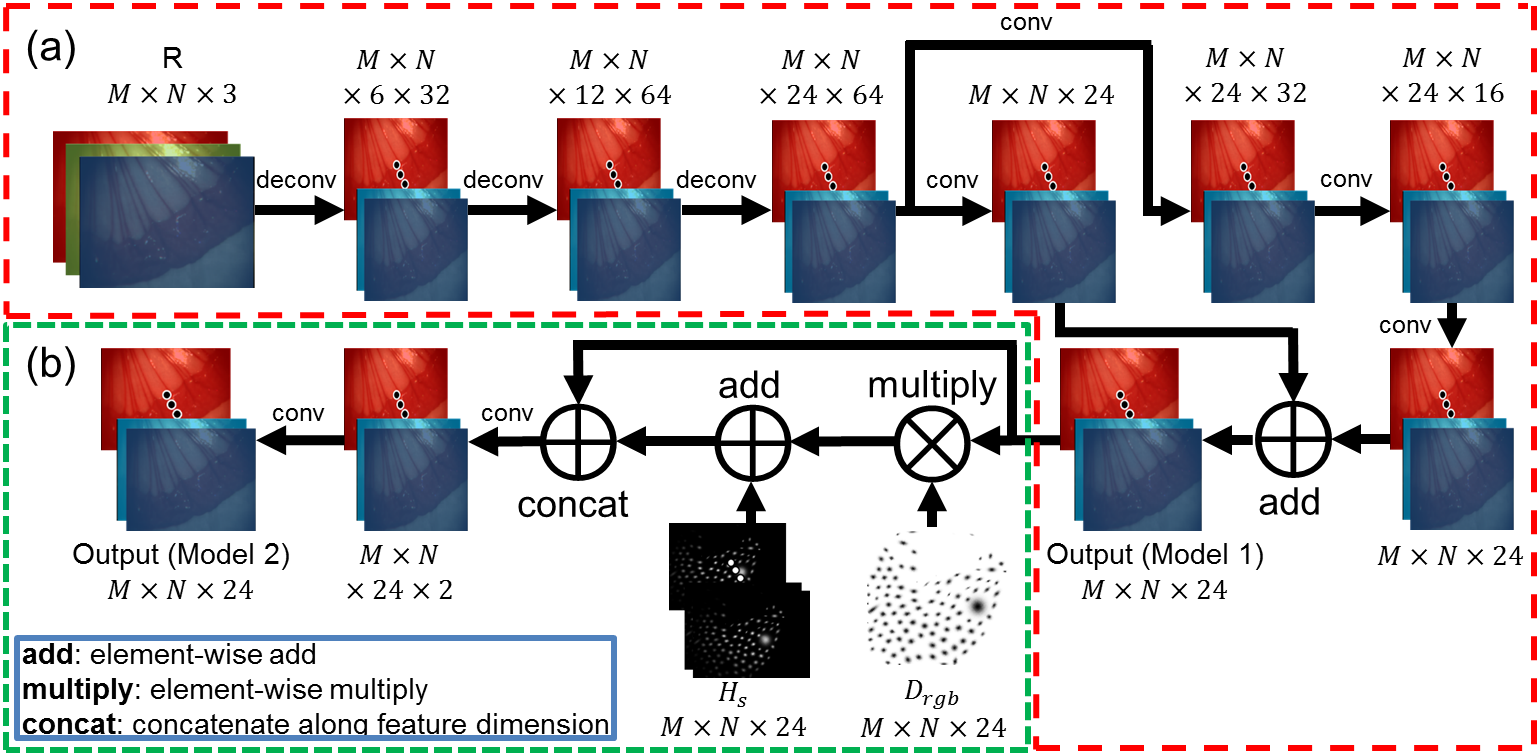}
\caption{(a) Model 1 during prediction. (b) The “mergence” layers added in model 2.}
\label{fig:Fig2}
\end{figure}

Model 2 takes three inputs: an RGB image ($M\times N\times 3$), a density map ($M\times N$) indicating the locations where the HSI is collected, and a sparse stack ($M\times N\times 24$) containing the sparse HSI signal. Model 2 added a “merging stage” on top of Model 1 (Fig.\ref{fig:Fig2} (b)), where all inputs were integrated. The HSI data was concatenated with the element-wise product between the density map and the HSI stack recovered from RGB. The final spatially dense MSI stack was estimated following a convolution.

\subsubsection{Training and prediction.}
Choosing a training set that provides adequate prior knowledge is of great importance for accurate MSI recovery. In this work, MSI stacks collected \textit{in vivo} during animal trials have been used for training and testing. The stacks ($H$) were collected using an LCTF endoscopic imager \cite{Clancy2015}, and different spectra were registered to create spectrally matched stacks. The transmission spectrum ($h$) for an RGB camera (Thorlabs DCU223C) was utilized to generate the synthetic RGB images ($R$) from HSI stacks, with $R = h\ast H$. The density map ($D_{hsi}$) for the sparse HSI signal was produced using previous spot segmentation results, with Gaussian distribution ($max = 1$) filled at each spot location (where the sparse spectral signal comes from). The density map for the RGB image $D_{rgb}$) was defined by $D_{rgb} = 1-D_{hsi}$. The sparse HSI stack ($H_{s}$) was the element-wise product between the density map and the HSI stack ($H_{s} = D_{rgb}\odot H$). 

To guarantee sufficient training samples, model 1 was trained on individual pixel spectral vectors instead of whole MSI stacks. In this case convolutions were applied along the spectral dimension, so that the trained network can be applied to inputs with arbitrary spatial dimensions. When training model 2, the network was initialized by the trained parameters from model 1. A two-stage training strategy was adopted instead of training directly from scratch: the parameters in the shared layers with model 1 were frozen while the “mergence” layers were updated; then all the parameters were updated until convergence. Both models were trained using Adam optimizer and $L2$-norm loss function. In prediction, RGB images were captured by the same camera, and the sparse HSI signal came from a HSI camera. Training and prediction were implemented using Tensorflow \cite{Abadi2016}. The prediction costs $~120 ms$ per frame on a PC (OS: Ubuntu 14.04; CPU: i7-3770; GPU: NVIDA GTX TITAN X). 

\section{Experimental Results}
\textit{In vivo} animal experimental data (MSI stack from 50 pig bowel, 21 rabbit uterus, 10 sheep uterus) were used to train and validate both models. By mixing the data from different sources and data augmentation, a 5-fold leave one-out cross-validation (LOOCV) was applied on a dataset containing 243 MSI stacks; each fold contained 200 MSI stacks for training and the remaining for testing.

\begin{figure}
\includegraphics[width=1\textwidth]{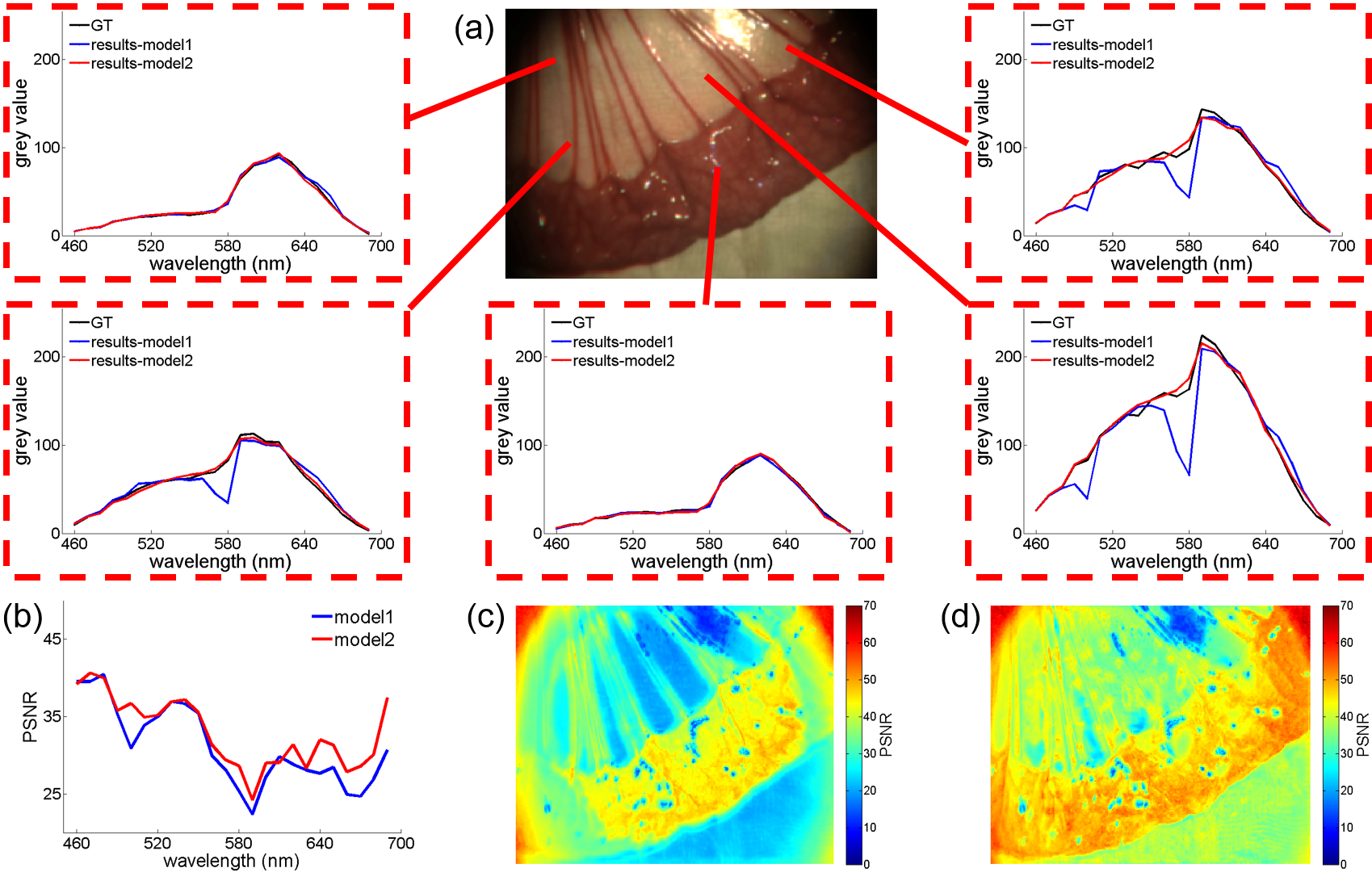}
\caption{(a) The RGB image and the estimated MSI (model 1: blue; model 2: red) vs. ground truth (black) from 5 locations. (b) PSNR along different wavelengths (model 1: blue; model 2: red) in LOOCV. PSNR map for model 1 (c) and model 2 (d) regarding the same sample (a).}
\label{fig:Fig3}
\end{figure}

Given the ground truth the peak signal-to-noise ratio ($PSNR = 20\lg(\frac{255}{MSE})$, MSE: mean square error) was adopted to evaluate the performances of both models. In the validation on average model 2 demonstrated significantly higher PSNR ($\approx 30.4$) compared with model 1 ($\approx 28.5$). The average PSNR on different wavelengths are shown in Fig.\ref{fig:Fig3} (b). In order to intuitively show the difference between two models, the estimated multi-spectral signals from 5 points, randomly chosen from representative areas in one pig bowel image, are compared (Fig.\ref{fig:Fig3} (a)). Although model 1 provided an estimation that generally fitted the ground truth, it suffered from large errors at some wavelengths. To the contrary model 2 provided improved accuracy over the entire spectral range. The pixel level PSNR maps for the estimated MSI regarding the same image are shown in Fig.\ref{fig:Fig3} (c, d). Excluding the saturation area, the minimum and mean PSNR are $13.1$ and $34.7$ for model 1, and $14.1$ and $42.0$ for model2.

Evaluation of transfer learning results is of great importance on machine learning problems, especially the clinical ones, where high model generalization capability is required. Thus, we also trained our models on data from different sources and tested them on each other. Table 1 lists the transfer learning results on two models, showing the importance of integrating the sparse HSI signal for more accurate MSI estimation.

\begin{table}[h!]
\caption{The transfer learning results (indicated by PSNR) with model 1 (left) and model 2 (right). Both models were trained and tested on 4 datasets: pig bowel (PB), rabbit uterus (RU), sheep uterus (SU), and RU+SU. 5-fold cross-validation was applied to calculate the PSNR when training and testing on data from the same source.}
\begin{tabular}{m{5em}|>{\centering}m{2.7em}|>{\centering}m{2.7em}|>{\centering}m{2.7em}|>{\centering}m{3.5em}} 
 \hline
\backslashbox{Train}{Test}  & PB & RU & RU & RU+SU \tabularnewline \hline
PB & 25.89 & 30.68 & 29.82 & 28.36 \tabularnewline \hline
RU & 25.07 & 33.24 & 32.34 & 32.59 \tabularnewline \hline
SU & 24.70 & 32.56 & 32.77 & 32.39 \tabularnewline \hline
RU+SU & 25.08 & 32.96 & 31.83 & 32.96 \tabularnewline \hline
\end{tabular}
\quad
\begin{tabular}{m{5em}|>{\centering}m{2.7em}|>{\centering}m{2.7em}|>{\centering}m{2.7em}|>{\centering}m{3.5em}} 
 \hline
\backslashbox{Train}{Test}  & PB & RU & RU & RU+SU \tabularnewline \hline
PB & 28.26 & 31.00 & 29.60 & 30.55 \tabularnewline \hline
RU & 26.95 & 34.22 & 33.00 & 33.83 \tabularnewline \hline
SU & 26.54 & 32.97 & 33.25 & 33.06 \tabularnewline \hline
RU+SU & 26.84 & 34.15 & 33.53 & 33.95 \tabularnewline \hline
\end{tabular}
\end{table}

\begin{figure}
\includegraphics[width=1\textwidth]{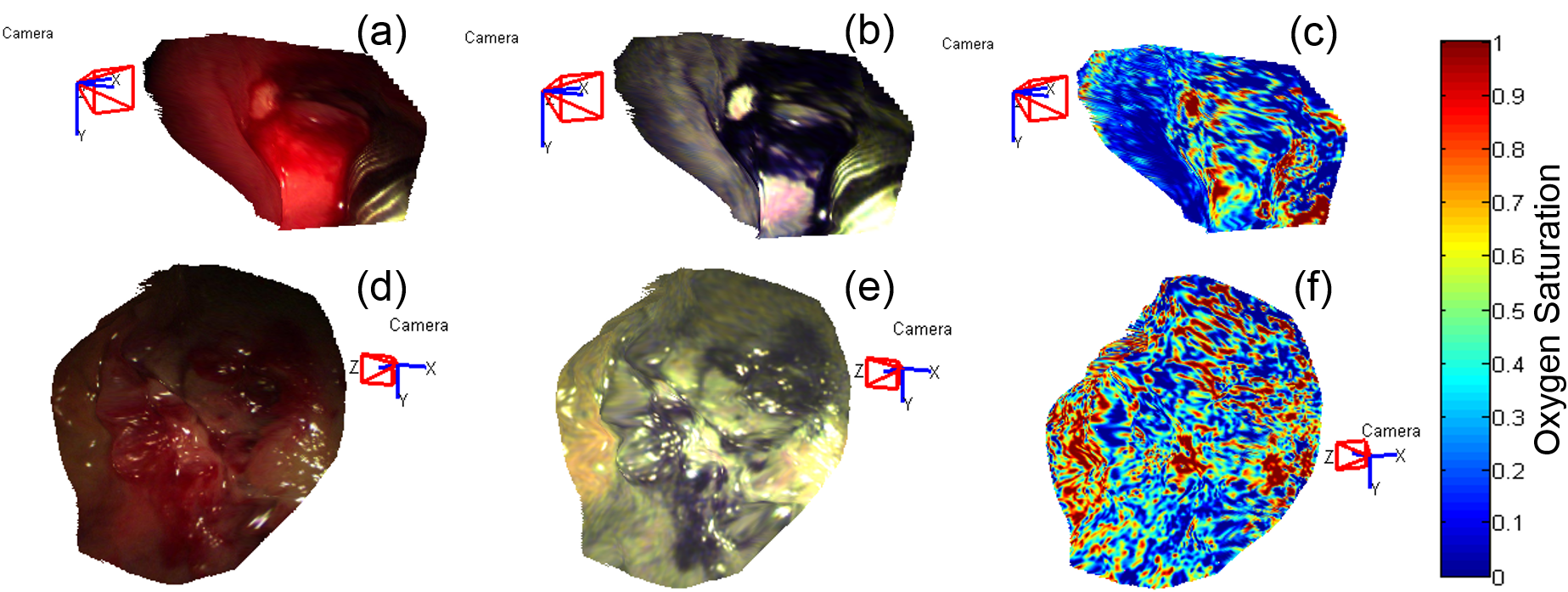}
\caption{Example human (a) \textit{in vivo} and (d) \textit{ex vivo} reconstructed tissue surface. (b, e) synthetic narrow band images and (c, f) synthetic oxygen saturation maps overlayed on these surfaces.}
\label{fig:Fig4}
\end{figure}

Acquiring the MSI stack is a prerequisite for imaging modalities like oxygen saturation and narrow band imaging, which could provide information to aid diagnosis and surgical navigation. As image examples, the oxygen saturation and narrow band images estimated from the MSI stacks, are overlaid onto the 3D reconstructed surfaces from \textit{ex vivo}/ \textit{in vivo} human experiments (Fig.\ref{fig:Fig4}).

\section{Discussion and Conclusion}
We have proposed a system capable of reconstructing tissue surface shape and re-covering dense multispectral signals. The implementation of interleaved SL and WL imaging provided WL views with shape and texture information that could be extracted for further applications e.g., object tracking or visual servoing, to benefit MIS and robotic surgery. The SfM pipeline can be further updated to state-of-the-art algorithms in future work. A near real-time ($\approx 8$ FPS) algorithm has been proposed to recover dense pixel-level multispectral signals. The accuracy and robustness of this algorithm have been demonstrated statistically and intuitively using different experimental results, \textit{in vivo} and \textit{ex vivo} from animal and human studies. oxygen saturation and narrow band imaging, that can be derived from the estimated MSI stack, were shown. The performance should be validated by further experiments on human tissue, especially abnormal structures like tumors and polyps. We believe the “super-spectral-resolution” algorithm can also benefit other general HSI acquisition modalities to greatly reduce scanning time with little compromise in performance.

\subsubsection{Ethics statement.}
The ethics approval for human study was covered by Central London Research Ethics Committee (reference No. $10/H0718/55$), animal study was conducted under UK Home Office license (reference No. $70/24843$, $70/7508$, $70/6927$, $8012639$).

\end{document}